\documentclass[final, 3p, times]{elsarticle}

\usepackage{lineno,hyperref}
\modulolinenumbers[5]

\journal{Information Processing and Management}

\usepackage{amsmath}
\usepackage{array}
\usepackage{graphicx}
\usepackage{subfloat}
\usepackage{multirow}
\usepackage{subfig}
\usepackage{makecell}
\usepackage{times}
\usepackage{latexsym}
\usepackage{booktabs} 
\usepackage{multirow}
\usepackage{url}
\usepackage{amsmath}
\usepackage{bm}
\usepackage{amsfonts}
\usepackage{graphicx}
\usepackage{diagbox}
\usepackage{xspace}
\usepackage{courier}
\usepackage{mathrsfs}
\usepackage{float}
\usepackage{framed} 
\usepackage{color}
\usepackage{amsfonts}
\usepackage{amssymb}
\usepackage{diagbox}
\usepackage{subfig}
\usepackage{array}
\hypersetup{hidelinks}
\usepackage[normalem]{ulem}
\usepackage{bm}
\usepackage{makecell}
\usepackage{multirow}
\usepackage{url}
\usepackage{graphicx}
\usepackage{appendix}
\usepackage{float}
\usepackage{color}
\usepackage{subfig}
\usepackage{array}
\usepackage{color}
\useunder{\uline}{\ul}{}
\usepackage{amsmath}
\usepackage{array}
\usepackage{graphicx}
\usepackage{subfloat}
\usepackage{multirow}
\usepackage{subfig}
\hypersetup{hidelinks}
\biboptions{sort&compress} 
 
\usepackage[labelfont=bf]{caption}  
 
\makeatletter 
\def\@makecaption#1#2{%
	\vskip\abovecaptionskip 
	\sbox\@tempboxa{#1 #2}%
	{\bfseries #1} #2\par 
	\vskip\belowcaptionskip} 
\makeatother 

\usepackage{color}
\definecolor{REVISE}{RGB}{0, 0, 255}

\begin{document}
	
	\begin{frontmatter}
		
		\title{DsMtGCN: A Direction-sensitive Multi-task framework for Knowledge Graph Completion}

		\author[mymainaddress]{Jining~Wang}
		
		\author[mymainaddress]{Chuan~Chen\corref{mycorrespondingauthor}}
		\cortext[mycorrespondingauthor]{Corresponding author}
		\ead{chenchuan@mail.sysu.edu.cn}
		
		\author[mymainaddress]{Zibin~Zheng}
		
		\author[mymainaddress]{Yuren~Zhou}
		
		\address[mymainaddress]{School of Computer Science and Engineering, Sun Yat-sen University, Guangzhou, 510275, China}
		
		\begin{abstract}
			To solve the inherent incompleteness of knowledge graphs (KGs), numbers of knowledge graph completion (KGC) models have been proposed to predict missing links from known triples. Among those, several works have achieved more advanced results via exploiting the structure information on KGs with Graph Convolutional Networks (GCN). However, we observe that entity embeddings aggregated from neighbors in different directions are just simply averaged to complete single-tasks by existing GCN based models, ignoring the specific requirements of forward and backward sub-tasks. In this paper, we propose a Direction-sensitive Multi-task GCN (DsMtGCN) to make full use of the direction information, the multi-head self-attention is applied to specifically combine embeddings in different directions based on various entities and sub-tasks, the geometric constraints are imposed to adjust the distribution of embeddings, and the traditional binary cross-entropy loss is modified to reflect the triple uncertainty. Moreover, the competitive experiments results on several benchmark datasets verify the effectiveness of our model.
		\end{abstract}
		
		
	\end{frontmatter}
	
	\section{Introduction}
	
	KGs are multi-relational directed graphs with entities as nodes and relations as edges, containing a large amount of structured knowledge in the form of (head entity, relation, tail entity) triplets, they are widely applied in question answering \cite{qatask}, dialog systems \cite{dstask}, web search \cite{wstask} and recommendation system \cite{rstask}. However, due to the limitation of available resources, it is impractical to store all facts in KGs, which leads to the incompleteness \cite{incompleteness}, and the algorithms of KGC are required to solve the problem.
	
	\begin{figure*}[!t]
		\centering
		\subfloat[]{\includegraphics[width=0.56\hsize]{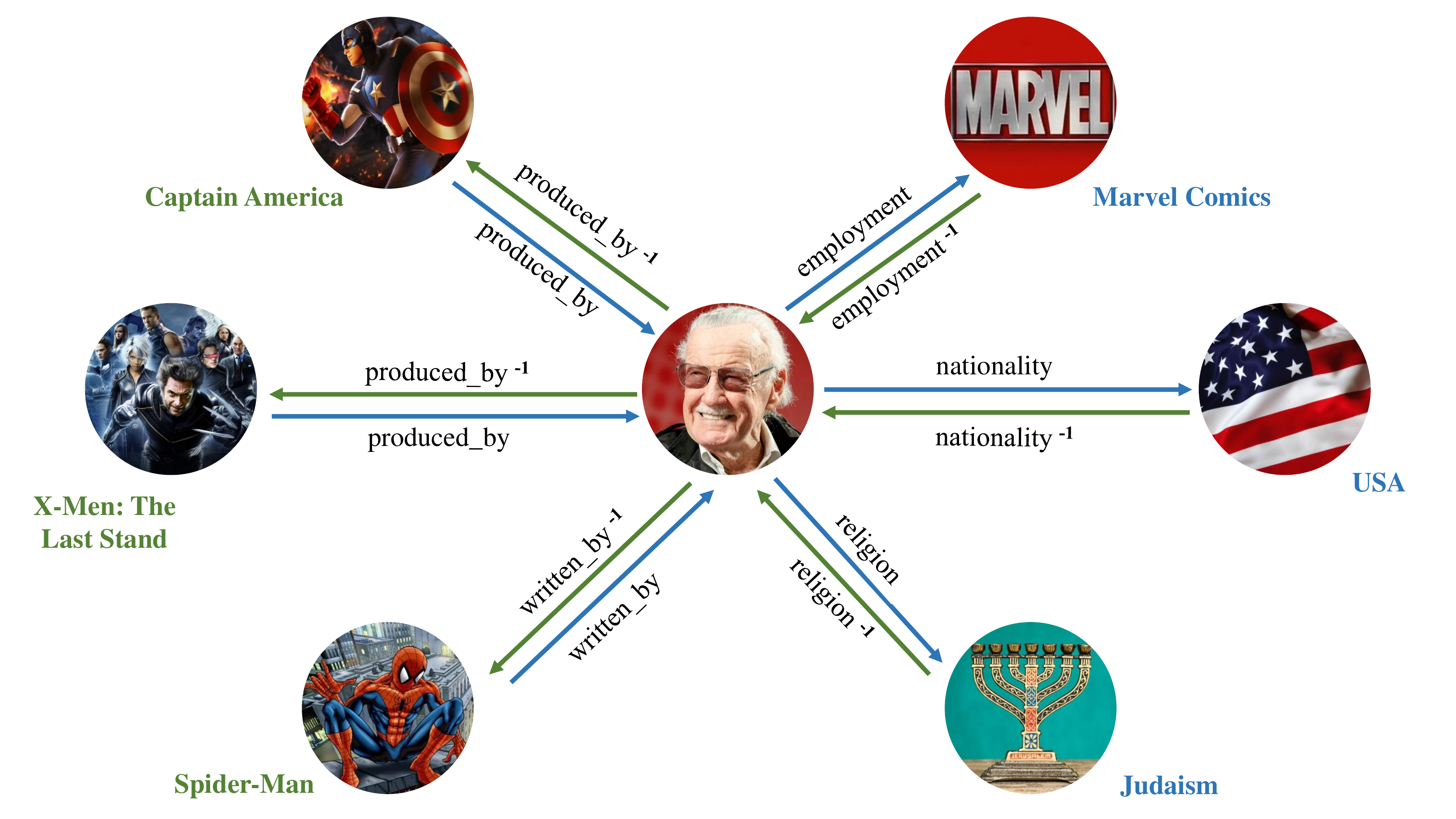}}  
		\subfloat[]{\includegraphics[width=0.44\hsize]{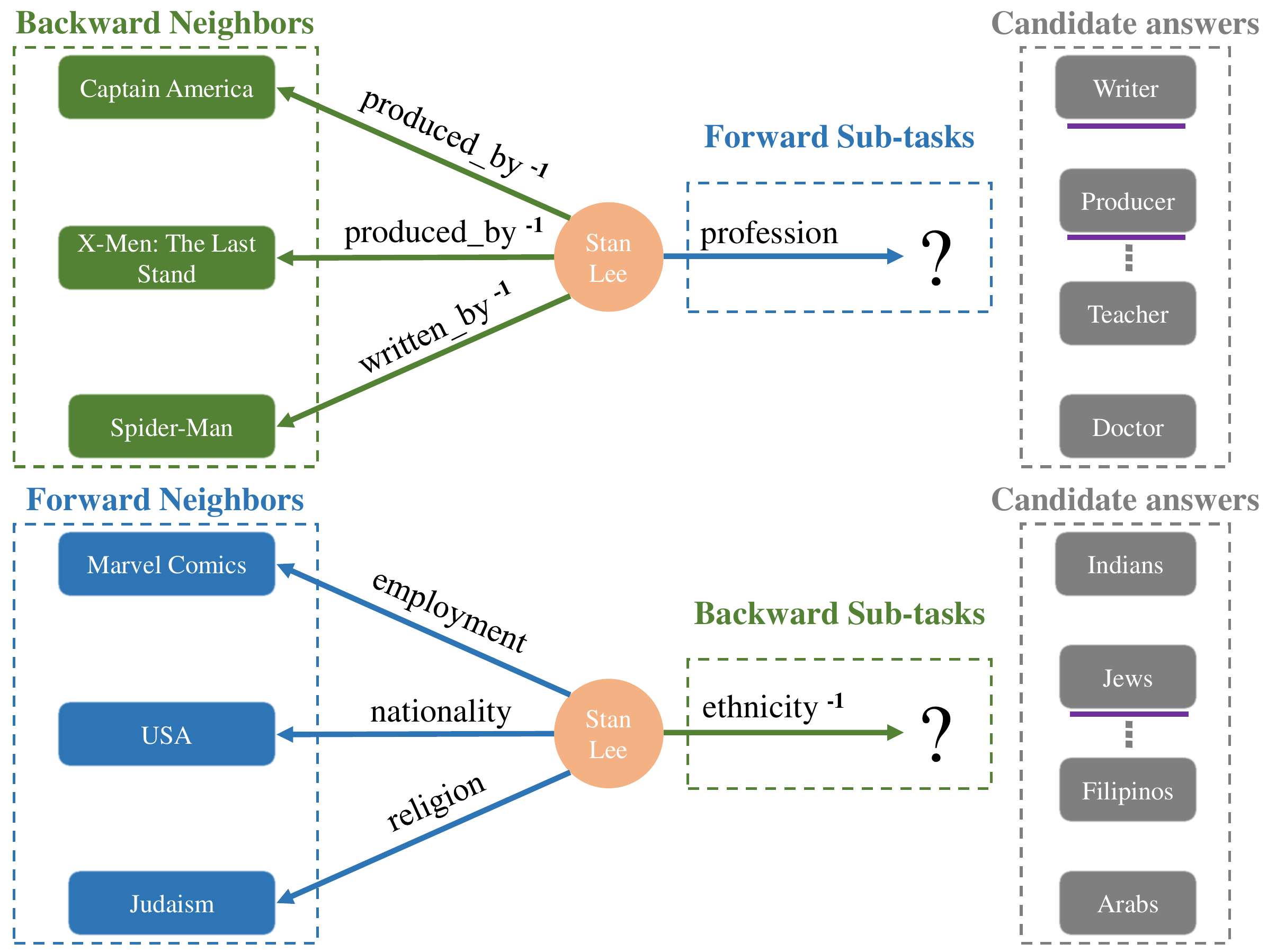}}
		\caption{An example of different sub-tasks' preferences for neighbors in forward and backward directions. (a) Knowledge graph around Stan Lee extracted from FB15k-237. Blue arrows indicate original relations in forward direction, while green ones mean inverse relations in backward direction written as $r^{-1}$. (b) The forward and backward sub-tasks about Stan Lee in FB15k-237, neighbors in specific directions which are critical to making decisions are listed on the left, correct answers are underlined in purple on the right.}
		\label{example}
	\end{figure*}
	
	There are a lot of researches focusing on KGC or link prediction tasks aiming to infer missing facts automatically based on known facts. Pioneering additive models \cite{transe, transh, transr} take the transformation from head entities to tail entities as a translation problem, while multiplicative models \cite{rescal, distmult, complex, quate} try to measure the plausibility of unknown triplets by applying proper semantic similarity-based score function. Benefiting from the development of neural networks, several works concentrate on the deeper nonlinear interactions among entities and relations with innovative model structures \cite{sme, nam, conve, convkb, convr, interacte}. 
	
	Furthermore, some recent studies introduce GCN to take the structure information into consideration by aggregating neighborhood information \cite{rgcn, sacn, vrgcn, compgcn}, which brings significant improvement. Despite the high-performance of them, they fail to utilize direction information implied in different neighbors while merging them, which is important for making reasonable predictions. As shown in Fig.~\ref{example}, there exists original and inverse edges (relations) in KGs, according to the direction of them, the link prediction tasks can be divided into forward and backward sub-tasks, neighbors can also be grouped into forward and backward neighbors, and sub-tasks in different directions always have diverse preferences for neighbors. For example, while dealing with forward sub-task \textit{(Stan Lee, profession, ?)}, it can be solved from backward neighbors including \textit{Spider-Man, Iron Man} and \textit{Captain America}; on the other hand, the answer for query \textit{(Stan Lee, ethnicity$^{-1}$, ?)} can be derived from \textit{Judaism}, which is a forward neighbor of \textit{Stan Lee}. However, existing studies can't model such preferences as they just simply average neighbor embeddings in different directions, which inspires our work.
	
	On the basis of above observations, we propose a novel model called DsMtGCN in this paper. Concretely, we design a multi-task framework to complete forward and backward prediction sub-tasks respectively to merge the direction information with differential partiality. In consideration of the differences among entities and sub-tasks, we use multi-head self-attention at entity level to combine embeddings aggregated from neighbors in various directions under forward and backward sub-tasks. Inspired by multi-view learning, we believe that the embeddings learnt by neighbors in different directions can be seen as two views obtained from two sides, and we impose distance and conicity constraints for embeddings to satisfy the principle of consensus and difference. Besides, we find that the difference on number of possible tail entities leads to triple uncertainty. For example, if the task is \textit{(Stan Lee, profession, ?)}, then there are numbers of correct answers and larger triple uncertainty. On the other hand, if the query is \textit{(Stan Lee, nationality, ?)}, the candidate tail entity is unique with small triple uncertainty. For this reason, we redefine label smoothing for link prediction tasks to fully considers the uncertainty while training.  
	
	In summary, our contributions in this paper are the following:
	
	\begin{itemize}
		\item We propose a multi-task model called DsMtGCN to utilize ignored direction information by considering the diverse direction preferences of forward and backward sub-tasks, which is the first study in this field to the best of our knowledge.
		\item For embeddings learned in different directions, we apply multi-head self-attention to extract direction information specifically at the entity level under different sub-tasks, and adopt the geometric constraints to adjust the embedding distribution.
		\item We propose an improved label smoothing under the multi-label classification scenario to reflect the triple uncertainty, which further increase the performance of our model.
		\item By analyzing experimental results of DsMtGCN on FB15k-237 and WN18RR datasets, we demonstrate the effectiveness of our proposed method.
	\end{itemize}
	
	\section{Related work}
	\label{rw}
	
	In this part, we will introduce related work from two aspects including link prediction and multi-task learning, they will be described in detail below.
	
	\subsection{Link Prediction}
	
	Link prediction or knowledge graph completion aims to predict missing triples by scoring candidate triples with embeddings of entities and relations leaned by well-designed models. Some non-neural approaches were first proposed in this field, they usually take relations as transfer vectors to convert head entities to tail entities, different models differ in score functions and representation spaces. For example, TransE\cite{transe}, TransH\cite{transh}, and TransR\cite{transr} are known as typical additive models, since they project entities and relations to vector space, then optimize the distance-based score function with the goal of making the head entity added by relation as close as possible to the tail entity. On the other hand, multiplicative models use semantic similarity-based score function to measure the plausibility of given triplets, enabling them to maintain good expression ability when the scale of the knowledge graph grows. Specifically, RESCAL\cite{rescal} and DistMult\cite{distmult} introduce bilinear operation in point-wise space, while Complex\cite{complex} and QuatE\cite{quate} represent embeddings in complex-valued space and four dimensional space respectively to capture complicated patterns like composition and antisymmetry of relations.
	
	With the wide application of neural networks in various fields, several neural network based models have been proposed to improve expression ability by capturing more complex interactions among triples. Considering that the simplicity and expressiveness of Multi-Layer Perception (MLP) are suitable for enhancing score functions, NTN\cite{ntn} and SME\cite{sme} explore nonlinear modeling with a fully-connected layer and activation function, while NAM\cite{nam} adopts deeper hidden layers and utilizes information from the hidden encoding. Convolutional Neural Networks (CNN) have also been employed to improve expression ability by learning deeper features with fewer shared parameters. For instance, ConvE\cite{conve} applies 2D convolution over a two-dimensional matrix reshaped from stacked head entities and relations to model the interactions among them, Conv-KB\cite{convkb} adopts 1D filters on the same dimension of entities and relations directly to keep better transitional characteristic, ConvR\cite{convr} replaces global filters used by ConvE with relation-specific ones, InteractE\cite{interacte} proposes several skills including feature permutation, checkered reshaping as well as circular convolution to maximize the number of heterogeneous and homogeneous interactions between entity and relation features. The information used by most methods is limited to the triple level as mentioned above, in order to make better decisions, RSN\cite{rsn} tries to capture long-term path dependence with Recurrent Neural Networks (RNN), while transformer-based models like CoKE\cite{coke} and KG-BERT\cite{kgbert} choose to mine contextual information from knowledge graphs. 
	
	Although knowledge graphs are fundamentally multi-relational graphs, the aforementioned methods, which are more suitable for Euclidean data, tend to overlook the structural information inherent in them. To make up for it, recent works introduce Graph Neural Networks (GNN) for graph context modeling under the encoder-decoder framework. R-GCN\cite{rgcn} proposes relation specific filters to take different relations into account while modeling the message passing on knowledge graphs. SACN\cite{sacn} learns adaptive weights of adjacent nodes with same relation instead of taking the neighborhood of each entity equally, so that node structure, node attributes and relation types can contribute to the performance gains. VR-GCN\cite{vrgcn} explicitly learns the representation of entities and relations to make full use of relation information. CompGCN\cite{compgcn} introduces entity-relation composition operations while aggregating neighbors, which is proven to generalize previous GCN-based models including KipfGCN\cite{kipfgcn}, R-GCN\cite{rgcn}, D-GCN\cite{dgcn} and SACN\cite{sacn}. KBGAT\cite{kbgat} claims to surpass others by learning graph attention based embeddings, but its results have been questioned by another study\cite{reevaluation}. Nevertheless, none of above studies suppose that neighbors in different directions should be specifically focused under different sub-tasks, which inspires our work.
	
	\subsection{Multi-task Learning}
	
	Multi-task Learning (MTL) is a specially designed network structure to leverage information across tasks instead of learning each task in isolation \cite{mtlreview}. Benefiting from its simplicity and efficiency, MTL has contributed to computer vision \cite{tcdcn, cross-stitch}, natural language processing \cite{mtdnn}, recommendation system \cite{mkr} and so on. 
	
	In the field of link prediction, existing MTL models mainly focus on introducing additional tasks or datasets. Specifically, SENN \cite{senn} utilizes relation prediction to promote link prediction, TransMTL \cite{transmtl} learns embeddings of multiple knowledge graphs simultaneously to mine shared structure patterns among them. However, the aforementioned methods bring much extra computational overhead, and it's not easy to find suitable sub-tasks or external datasets. In this paper, we define sub-tasks by decomposing existing tasks, which reduces the number of parameters and avoids introducing extra noise.
	
	\section{Background and Definition}
	
	In this section, we aim to introduce several important concepts and notations related to our work.
	
	\subsection{Link prediction tasks}
	We denote KGs as $\mathcal{G}=\{(h, r, t)  \mid  (h, r, t) \in \mathcal{T}, h, t \in \mathcal{E}, r \in \mathcal{R}\}$, where $h, t$ are head and tail entity, $r$ represents the relation connecting $h$ and $t$, $\mathcal{T, E, R}$ denote the set of triples, entities, and relations. The train, valid and test set of triples are denoted as $\mathcal{T}^{train}$, $\mathcal{T}^{valid}$ and $\mathcal{T}^{test}$ respectively.
	
	Given $(h, r)$, the link prediction task aims to predict missing $t$ on $\mathcal{G}$. Let $e_h, e_r, e_t \in R^d$ be corresponding embeddings with dimension $d$, a score function defined as $\psi: {R}^{d} \times {R}^{d} \times {R}^{d} \rightarrow {R}$ takes $e_h, e_r, e_t$ as input and outputs a score measuring the plausibility of triplet $(h, r, t)$, then $t$ with the highest score is selected as the prediction result.
	
	\subsection{Link prediction sub-tasks}
	
	Following  \cite{bidirection}, the information in KGs is required to flow in backward and self-loop directions besides forward direction. For triple $(h, r, t)$, we generate its backward version $(t, r^{-1}, h)$ and self-loop version $(h, r^0, h)$ to augment original data. After that, the new relation set is $\mathcal{R}' = \mathcal{R} \cup \mathcal{R}^{-1} \cup \{r^0\}$, where $\mathcal{R}^{-1} = \{r^{-1}  \mid  r \in \mathcal{R}\}$; the new triple set is $\mathcal{T}' = \mathcal{T} \cup \mathcal{T}^{-1} \cup \mathcal{T}^{0}$, where $\mathcal{T}^{-1} = \{(t, r^{-1}, h)  \mid  (h, r, t) \in \mathcal{T}\}$, $\mathcal{T}^{0} = \{(h, r^{0}, h)  \mid  h \in \mathcal{E}\}$, the prediction tasks can be divided into forward sub-tasks like $(h, r, ?)$ and backward sub-tasks like $(h, r^{-1}, ?)$. 
	
	For one entity $h$, its neighbors $\mathcal{N}_h$ can be grouped into forward neighbors $\mathcal{N}^F_h$ and backward neighbors $\mathcal{N}^B_h$, where ${N}^F_h = \{(r, t)  \mid  (h,r,t) \in \mathcal{T}\}$, ${N}^B_h = \{(r^{-1}, t)  \mid  (h,r^{-1},t) \in \mathcal{T}^{-1}\}$, and they are considered to contribute differently to the two sub-tasks. Furthermore, the embedding representation learned from $\mathcal{N}^F_h$ and $\mathcal{N}^B_h$ are denoted as $e^f_h$ and $e^b_h$ respectively, the set of $e^f_h$ and $e^b_h$ can be rewritten in the matrix form as $E^f, E^b \in R^{|\mathcal{E}| \times d}$.
	
	\section{Model}
	\label{md}
	
	\begin{figure*}[!t]
		\centering
		\includegraphics[width=1.0\hsize]{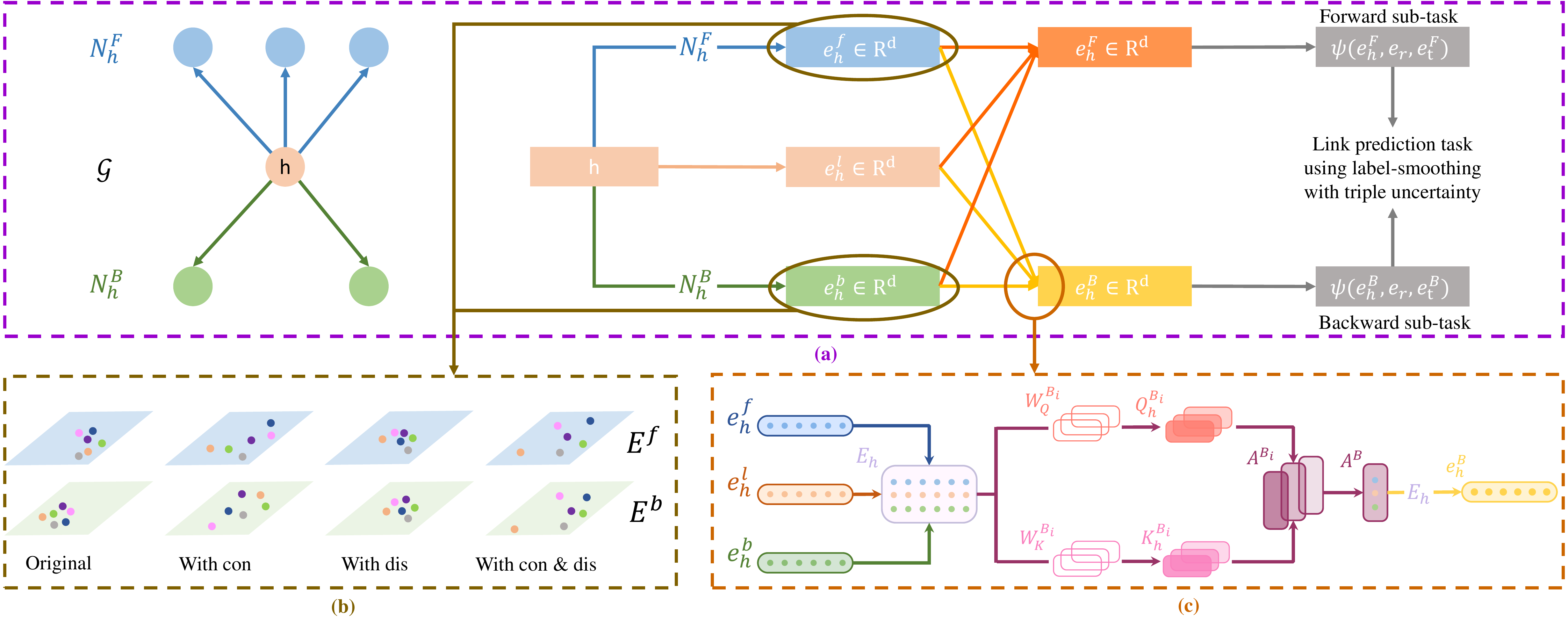}
		\caption{The structure of DsMtGCN and modules it contains. (a) An overview of our DsMtGCN, the input data are shown on the left. (b) Influence of different geometric constraints, the blue and green plane indicate the space in which $E^f$ and $E^b$ are located, colored points correspond to embeddings of different entities. (c) The illustration explaining how the multi-head self-attention mechanism works to get $e^B_h$ for backward prediction sub-task.}
		\label{model}
	\end{figure*}
	
	In this part, we will describe the structure of DsMtGCN. Overall, the essential idea of our work is to integrates neighbor information in specific directions at entity level based on sub-tasks so that the direction information can be fully used, Fig.~\ref{model}(a) provides an intuitive explanation to this point. In what follows,we will introduce each module of DsMtGCN and the joint using of them in detail.
	
	\subsection{Message Passing}
	
	Inspired by CompGCN, we use entity-relation composition operation $\phi$ and message passing function $M$ to model the information flow from neighbors in different directions. 
	
	Specifically, the features of entity $h$ and relation $r$ are initialized as $v_h, v_r\in R^{d_i}$. Given one known triple $(h, r, t)$, the operation $\phi$ is applied to get the distributed representation of $(r, t)$, which is implemented in two ways including \textbf{Multiplication (Mult)} \cite{distmult} and \textbf{Circular-correlation (Corr)} \cite{hole} defined as follows, where $\circ$ denotes Hadamard product and $\star$ means compressing the tensor product.
	
	\begin{equation}
		\textbf{Mult:}\: \phi(r, t) = v_r \circ v_t
	\end{equation}
	
	\begin{equation}
		\textbf{Corr:}\: \phi(r, t) = v_r \star v_t
	\end{equation}
	
	The representations are then aggregated by $M$, where $W^F,W^B,$ $W^L \in R^{d_i \times d}$ are direction-specific weights.
	\begin{equation}
		e^f_h = M(N^F_h, W^F) = \sum_{(r, t) \in N^F_h} W^F \phi(r,t)
	\end{equation}
	\begin{equation}
		e^b_h = M(N^B_h, W^B) = \sum_{(r^{-1}, t) \in N^B_h}  W^B \phi(r^{-1},t)
	\end{equation}
	\begin{equation}
		e^l_h = M(\{(r^0,h)\}, W^L) =  W^L \phi(r^0,h)
	\end{equation}
	
	\subsection{Geometric constraints}
	
	In essence, $e^f_h$ and $e^b_h$ can be regarded as two views of $h$, which are expected to satisfy the principle of consensus and difference. Specifically, the consensus principle means $e^f_h$ and $e^b_h$ should be close to each other appropriately considering that they share similar semantics, the difference principle claims that embeddings of different entities from the same view should be distinguished as a whole. 
	
	For keeping the two principles, we impose distance constrains $dis(E^f, E^b)$ to bring them closer, the conicity geometric constrains $con(E^f)$ and $con(E^b)$ are used to increase the distinguishability of embeddings between different entities. The conicity term is originally defined as a geometric attribute \cite{conicity}, while it is regarded as an explicit constraint here. The formula of $dis$ and $con$ are shown as follows, where $E^f_i$ stands for the $i$-th row of $E^f$, $\Vert x-y \Vert_2$ and $cos(x, y)$ denote the Euclidean distance and cosine similarity.
	
	\begin{equation}
		dis(E^f, E^b) = \frac{1}{|\mathcal{E}|}\sum_{i = 1}^{|\mathcal{E}|} \Vert E^f_i-E^b_i \Vert_2
	\end{equation}
	
	\begin{equation}
		con(E^f) = \frac{1}{|\mathcal{E}|}\sum_{i = 1}^{|\mathcal{E}|} cos(E^f_i, \frac{\sum_{i = 1}^{|\mathcal{E}|}E^f_i}{|\mathcal{E}|})
	\end{equation}
	
	We illustrate the effect of constraints in Fig.~\ref{model}(b), the conicity constraints make it easier to distinguish points on the same plane, while the distance constraint can help to align points of the same color on different planes, the combination of them can achieve the superb result.
	
	\subsection{Attention mechanism}
	Considering the specific requirements for neighbor information in different directions under forward and backward sub-tasks, the multi-head self-attention mechanism \cite{need} at entity level is applied.
	
	For simplicity, we first expound how to calculate $e^F_h$ for the forward sub-tasks with single head. Suppose the dimension of attention vectors is $d_a$, given stacked embedding $E_h = [e_h^f; e_h^l; e_h^b] \in R^{3 \times d}$, we obtain the query vector $Q_h^F \in R^{3 \times d_a}$ and the key vector $K_h^F \in R^{3 \times d_a}$ as follows, where $W^F_Q, W^F_K \in R^{d \times d_a}$ are task-specific weights.
	\begin{equation}
		Q_h^F = E_h W^F_Q,	K_h^F = E_h W^F_K
	\end{equation}
	
	Then the scaled dot-product result $S^F \in R^{3 \times 3}$ can be formulated as:
	\begin{equation}
		S^F = \frac{Q_h^F {K_h^F}^T}{\sqrt{d_{a}}}
	\end{equation}
	
	The final attention $A^F \in R^{3}$ can be obtained after softmax and mean-pooling ($mp$), the $mp$ operator aims to save mean value of each column of the matrix to be processed, note that the sum of $A^F$ equals 1, the $m$-th element is denoted as $A^F_m$:
	
	\begin{equation}
		A^F = mp(softmax(S^F))
	\end{equation}
	
	\begin{equation}
		A^F_m = \frac{1}{3} \sum_{j = 1}^3 \frac{exp(S^F_{jm})}{\sum_{k = 1}^3 exp(S^F_{jk})}
	\end{equation}
	
	To get richer latent information and enhance the stability of results, we can use more heads to compute $A^F$. If the number of heads is $n_h$, the updated formula is written down below, where ${Q_h^{F_i}} = E_h {W_Q^{F_i}}$, ${K_h^{F_i}} = E_h {W^{F_i}_K}$:
	\begin{equation}
		A^F  =  \frac{\sum_{i=1}^{n_h} A^{F_i}}{n_h}  =  \frac{\sum_{i=1}^{n_h} mp(softmax(\frac{{Q_h^{F_i}} {K_h^{F_i}}^T}{\sqrt{d_{a}}}))}{n_h}
	\end{equation}
	
	In the end, $e_h^F \in R^d$ is the weighted sum of $e_h^f, e_h^l, e_h^b$ as follows, and $e_h^B$ can be calculated with similar process as shown in Fig.~\ref{model}(c).
	\begin{equation}
		e_h^F = A^F_1 e_h^f + A^F_2 e_h^l + A^F_3 e_h^b = A^FE_h
	\end{equation}
	
	\subsection{Loss function}
	
	In this work, we use ConvE to model the interaction between entities and relations. Specifically, the score function $\psi$ is defined as follows, the convolution operation is applied on the 2D reshaping of $e_h, e_r$ with filter $\omega$, and $g$ denotes the nonlinear activation function.
	\begin{equation}
		\psi(e_h, e_r, e_t) = g(vec(g([e_h; e_r] * \omega)) W) e_t
	\end{equation}
	
	Taking into account the different sub-tasks, one triple is scored as follows, where $\sigma$ is the sigmoid function mapping score to the interval of $(0,1)$, $\alpha$ is the indicator variable, whose value is 1 if current prediction task belongs to the forward prediction sub-tasks otherwise 0.
	\begin{equation}
		s(h, r, t)  =  \sigma(\alpha\psi(e_h^F, e_r, e_t^F) + (1-\alpha)\psi(e_h^B, e_r, e_t^B))
	\end{equation}
	
	For most previous works, the target of training is to minimize such standard binary cross-entropy loss for $i$-th triple, where $s_i$ is the predicted score, the label $t_i$ indicates the existence or non-existence of the triple, and $l$ controls the degree of label smoothing.
	
	\begin{equation}
		\mathcal{L}_{BCE}^i  =  \left\{\begin{array}{l} 
			-\log (\sigma(s_i))+\left(l-\frac{1}{|\mathcal{E}|}\right) * s_i, \text{if }t_i = 1\\
			-\log (\sigma(-s_i))-\frac{1}{|\mathcal{E}|} * s_i, \text{if }t_i = 0 
		\end{array}\right.
	\end{equation}
	
	To model the triple uncertainty, we propose a new label smoothing mechanism as follows to prevent the model from giving too low scores to potential targets, where $u_i = |\{t|(h_i, r_i, t) \in \mathcal{T}^{train}\}|$ ranging from 1 to $|\mathcal{E}|$ is the measurement of uncertainty of $i$-th triple, and $k$ controls the scale of adjustment:
	
	\begin{equation}
		\mathcal{L}_{TU}^i  =  \left\{\begin{array}{l}
			-\log (\sigma(s_i))+\left(l-\frac{u_i^k}{|\mathcal{E}|}\right) * s_i, \text{if }t_i = 1\\
			-\log (\sigma(-s_i))-\frac{u_i^k}{|\mathcal{E}|} * s_i, \text{if }t_i = 0 
		\end{array}\right.
	\end{equation}
	
	The final loss function $\mathcal{L}$ is defined as follows, where $\lambda_1$ and $\lambda_2$ are trade-off parameters:
	
	\begin{equation}
		\mathcal{L}  =  \mathcal{L}_{TU}  +  \mathcal{L}_{GC}
	\end{equation}
	
	\begin{equation}
		\mathcal{L}_{TU}  =  \frac{1}{|\mathcal{T}^{train}|}  \sum_{i=1}^{|\mathcal{T}^{train}|}  \mathcal{L}_{TU}^i
	\end{equation}
	
	\begin{equation}
		\mathcal{L}_{GC}  =  \lambda_1 dis(E^f , E^b)  +  \lambda_2 (con(E^f)  +  con(E^b))
	\end{equation}
	
	\section{Experiments}
	\label{ep}
	
	\begin{table}[!t]
		\centering
		\caption{Statistics of datasets.}
		\begin{tabular}{l|cc}
			\toprule
			\textbf{\textbf{Statistics}}                                  & \textbf{FB15k-237} & \textbf{WN18RR} \\ 
			\midrule
			$|\mathcal{E}|$  & 14,541    & 40,943 \\ 
			$|\mathcal{R}|$   & 237       & 11     \\ 
			$|\mathcal{T}^{train}|$	 & 272,115   & 86,835 \\ 
			$|\mathcal{T}^{valid}|$    & 17,535    & 3,034  \\ 
			$|\mathcal{T}^{test}|$       & 20,466    & 3,134  \\ 
			\bottomrule
		\end{tabular}
		\label{dataset}
	\end{table}
	
	In this section, we will demonstrate the effectiveness of DsMtGCN with numerous experiments, the experimental setting and results analysis will be introduced below.
	
	\subsection{Experimental Setting}
	
	\paragraph{Datesets}
	
	To confirm the performance of DsMtGCN in a fair way, FB15k-237 \cite{fb15k237} and WN18RR \cite{conve}, the two commonly used benchmarks, are adopted in our research. In particular, FB15k-237 provides general facts such as actors, companies and films, WN18RR contains semantic knowledge of lexicons like synonym and hyponym, both of them have their inverse relations removed to resolve the test leakage problem. Table \ref{dataset} shows the statistics details of the above datasets.
	
	\paragraph{Metrics}
	
	
	To evaluate the performance of different models on the link prediction task, several popular ranking-based metrics including mean reciprocal rank (MRR) and Hits@k (H@k) for k = 1, 3, 10 are reported, to avoid the unfairness caused by the model's tendency to output same scores for all triples, we adopt the random evaluation protocols \cite{reevaluation}. Suppose $r_i$ is the ranking of the target tail entity for $i$-th triple under filter setting, where valid candidates different from the target are filtered \cite{transe}, MRR is defined as the mean value of $\frac{1}{r_i}$ \cite{mrr}, while H@k measures the average proportion of $r_i$ less than k, the formulas are defined as follows, where $\mathcal{I}$ is the indicator function, whose value is 1 if the condition can be satisfied else 0.
	
	\begin{equation}
		MRR = \frac{1}{|\mathcal{T}^{test}|} \sum_{i=1}^{|\mathcal{T}^{test}|} \frac{1}{r_i}
	\end{equation}
	
	\begin{equation}
		Hits@k = \frac{1}{|\mathcal{T}^{test}|} \sum_{i=1}^{|\mathcal{T}^{test}|} \mathcal{I}(r_i <= k)
	\end{equation}
	
	\paragraph{Baselines}
	
	To illustrate the effectiveness of our model, the results of DsMtGCN are compared with following baselines.
	
	\begin{itemize}
		\item \textbf{Non-neural models:} Additive models including TransE \cite{transe}, Complex \cite{complex}, RotatE \cite{rotate}. Multiplicative models including CrossE \cite{crosse} and DistMult \cite{distmult}.
		\item \textbf{Neural network based models:} Models leveraging CNN or GCN to capture complex interactions between entities and relations. Typical models of the former include ConvE \cite{conve}, ConvKB \cite{convkb} and ConvR \cite{convr}, while R-GCN \cite{rgcn}, SACN \cite{sacn}, VR-GCN \cite{vrgcn}, KBGAT \cite{kbgat}, CompGCN \cite{compgcn} are in the latter case.
		\item \textbf{Multi-task models:} Due to the lack of results of SENN \cite{senn}, we only compare the TransMTL-E \cite{transmtl} model.
	\end{itemize}
	
	\paragraph{Implementations}
	
	We implement our model in PyTorch \cite{pytorch}, the Adam optimizer \cite{adam} is applied to speed up the training process. The optimal hyperparameters are determined by the best MRR evaluated on $\mathcal{T}^{valid}$. Concretely, we implement $\phi$ with Corr and Mult for FB15k-237 and WN18RR \cite{compgcn}, and other hyperparameters are confirmed by grid search. For example, the learning rate ranges from $1e-4$ to $1e-2$, the batch size is selected from $\{64, 128, 256, 512\}$, $d_i$ and $d$ are searched in $\{100, 200, 300, 400\}$, both $l$ and $k$ are within the range of 0 to 1. For FB15k-237, we set learning rate as 0.001, batch size as 128, $d_i$ as 100, $d$ as 200, $l$ as 0.2 and $k$ as 0.2. For WN18RR, we set learning rate as 0.0003, batch size as 256, $d_i$ as 400, $d$ as 200, $l$ as 0.1 and $k$ as 0.5.
	
	\subsection{Link Prediction Tasks}
	
	\begin{table*}[!t]
		\centering
		\caption{Link prediction results of DsMtGCN and other baselines. The performance of other models are taken from original or latest published papers, "-" denotes missing results, the best results are in bold. }
		\begin{tabular}{l|cccc|cccc}
			\toprule
			\multirow{2}{*}{\textbf{Model}}  & \multicolumn{4}{c|}{\textbf{FB15k-237}}                                                                                                                                               & \multicolumn{4}{c}{\textbf{WN18RR}}                                                                                                                                 \\
			&\textbf{MRR}                                & \textbf{H@1}                             & \textbf{H@3}                              & \textbf{H@10}                              &  \textbf{MRR}                       & \textbf{H@1}                             & \textbf{H@3}                              & \textbf{H@10}                    \\ \midrule
			TransE     & 0.294                              & -                                  & -                                  & 0.465                               & 0.226                     & -                                  & -                                  & 0.501                     \\
			Complex   & 0.247                              & 0.158                              & 0.275                              & 0.428                               & 0.440                       & 0.410                               & 0.460                               & 0.510                      \\
			RotatE     & 0.338                              & 0.241          & 0.375          & 0.533          & 0.476                     & 0.428                              & 0.492                              & 0.571                     \\
			CrossE     & 0.299                              & 0.211                              & 0.331                              & 0.474                               & -                         & -                                  & -                                  & -                         \\
			DistMult & 0.241                              & 0.155                              & 0.263                              & 0.419                               & 0.430                      & 0.390                               & 0.440                               & 0.490                      \\ \midrule
			ConvE       & 0.325                              & 0.237                              & 0.356                              & 0.501                               & 0.430                      & 0.400                               & 0.440                               & 0.520                      \\
			ConvKB     & 0.243                              & 0.155                              & 0.371                              & 0.421                               & 0.249                      & 0.057                              & 0.417                              & 0.524                     \\
			ConvR       & 0.350                              & 0.261                              & 0.385                              & 0.528                               & 0.475                     & 0.443                              & 0.489                              & 0.537                     \\ \midrule
			R-GCN        & 0.248                              & 0.151                              & -                                  & 0.417                               & -                         & -                                  & -                                  & -                         \\
			SACN         & 0.350                               & 0.261                              & 0.390                               & 0.540                                & 0.470                      & 0.430                               & 0.480                               & 0.540                      \\
			VR-GCN      & 0.248                              & 0.159                              & 0.272                              & 0.432                               & -                         & -                                  & -                                  & -                         \\
			KBGAT       & 0.157                              & -                                  & -                                  & 0.331                               & 0.412                     & -                                  & -                                  & \textbf{0.554}            \\
			CompGCN   & 0.355                              & 0.264                              & 0.390                               & 0.535                               & 0.479                     & 0.443                     & 0.494                     & 0.546                     \\ \midrule
			TransMTL-E     & 0.336                              & -                                  & -                                  & 0.526                               & 0.363                     & -                                  & -                                  & 0.541                     \\ \midrule \midrule
			\textbf{DsMtGCN}                    & \textbf{0.363} & \textbf{0.273} & \textbf{0.398} & \textbf{0.545} & \textbf{0.481} & \textbf{0.445} & \textbf{0.495} & 0.544 \\ 
			\bottomrule
		\end{tabular}
		\label{compare_result}
	\end{table*}
	
	We compare our DsMtGCN model with various baselines, the results are summarized in Table \ref{compare_result}, it can be observed that with the model structure becomes more complex, the overall effect improves, CNN based models surpass non-neural models, and GNN based models achieve better performance with the leverage of graph structures. Compared with all of the above models, DsMtGCN outperforms them under all metrics on FB15k-237 and 2 out of 4 metrics on WN18RR, indicating the strong competitiveness of our model.
	
	Compared with other GCN models like R-GCN, SACN and CompGCN, the improvement of our model can be attributed to the well designed multi-task framework, on the other hand, the superior performance over previous MTL models can be explained with the introduce of divided sub-tasks and structure information aggregated from neighbors, which demonstrates the effectiveness of our model. 
	
	We can also find several differences between different metrics and datasets from the above results. In particular, MR is not a proper metric as it always fails to keep pace with other metrics, a smaller MR doesn't means larger MRR and Hits@k, as the simple averaging it adopts will inevitably be affected by extreme values. Besides, the performance of overly complex models on WN18RR is limited\cite{reevaluation}, as it only contains 11 relations, the dataset is too simple to avoid over-fitting. In view of the complexity of knowledge graph in reality as well as the fairness of comparison, our follow-up experiments will pay more attention to the performance of metrics other than MR on FB15k-237.
	
	\subsection{Link Prediction Sub-tasks}
	
	\begin{table*}[!t]
		\centering
		\caption{Results on link prediction sub-tasks by relation category on FB15k-237, the forward and backward sub-tasks correspond to Tail and Head Prediction in some other works.}
		\begin{tabular}{ll|cc|cc|cc|cc|cc}
			\toprule
			\multicolumn{2}{c|}{\multirow{2}{*}{\textbf{\textbf{Task}}}}           & \multicolumn{2}{c|}{\textbf{ConvE}} & \multicolumn{2}{c|}{\textbf{InteractE}} & \multicolumn{2}{c|}{\textbf{SACN}} & \multicolumn{2}{c|}{\textbf{CompGCN}} & \multicolumn{2}{c}{\textbf{DsMtGCN}}       \\  
			\multicolumn{2}{c|}{}                                 & \textbf{MRR}         & \textbf{H@10}      & \textbf{MRR}           & \textbf{H@10}        & \textbf{MRR}         & \textbf{H@10}      & \textbf{MRR}      & \textbf{H@10}           & \textbf{MRR}            & \textbf{H@10}        \\ \midrule
			\multicolumn{1}{l|}{\multirow{4}{*}{\makecell[l]{Forward \\ Sub-tasks}}} & 1-1 & 0.193       & 0.385        & 0.386         & 0.547          & 0.422       & 0.547        & 0.457    & \textbf{0.604}    & \textbf{0.468} & 0.589          \\
			\multicolumn{1}{l|}{}                           & 1-N & 0.068       & 0.116        & 0.106         & 0.192          & 0.093       & 0.187        & 0.112    & 0.190             & \textbf{0.113} & \textbf{0.214} \\
			\multicolumn{1}{l|}{}                           & N-1 & 0.438       & 0.638        & 0.466         & 0.647          & 0.454       & 0.647        & 0.471    & 0.656             & \textbf{0.484} & \textbf{0.67}  \\
			\multicolumn{1}{l|}{}                           & N-N & 0.246       & 0.436        & 0.276         & 0.476          & 0.261       & 0.459        & 0.275    & 0.474             & \textbf{0.286} & \textbf{0.486} \\ \midrule
			\multicolumn{1}{l|}{\multirow{4}{*}{\makecell[l]{Backward \\ Sub-tasks}}} & 1-1 & 0.177       & 0.391        & 0.368         & 0.547          & 0.406       & 0.531        & 0.453    & \textbf{0.589}    & \textbf{0.458} & 0.583          \\
			\multicolumn{1}{l|}{}                           & 1-N & 0.756       & 0.867        & 0.777         & 0.881          & 0.771       & 0.875        & 0.779    & 0.885             & \textbf{0.787} & \textbf{0.888} \\
			\multicolumn{1}{l|}{}                           & N-1 & 0.049       & 0.09         & 0.074         & 0.141          & 0.068       & 0.139        & 0.076    & 0.151             & \textbf{0.084} & \textbf{0.155} \\
			\multicolumn{1}{l|}{}                           & N-N & 0.369       & 0.587        & 0.395         & 0.617          & 0.385       & 0.607        & 0.395    & 0.616             & \textbf{0.403} & \textbf{0.623} \\ \bottomrule
		\end{tabular}
		\label{relation_result}
	\end{table*}
	
	\begin{figure*}[!t]
		\centering
		\subfloat[]{\includegraphics[width=0.33\hsize, height = 0.25\hsize]{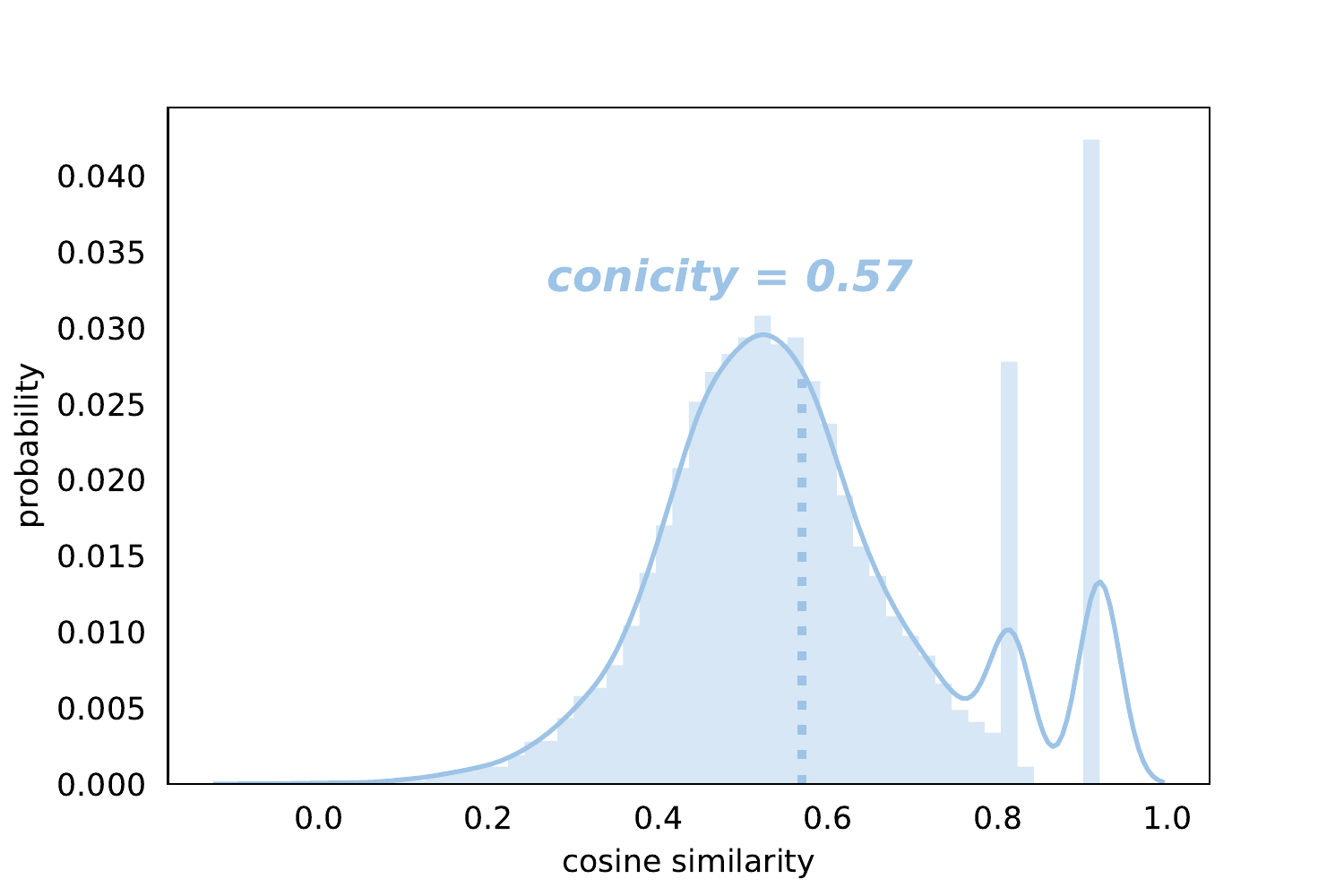}}  
		\subfloat[]{\includegraphics[width=0.33\hsize, height = 0.25\hsize]{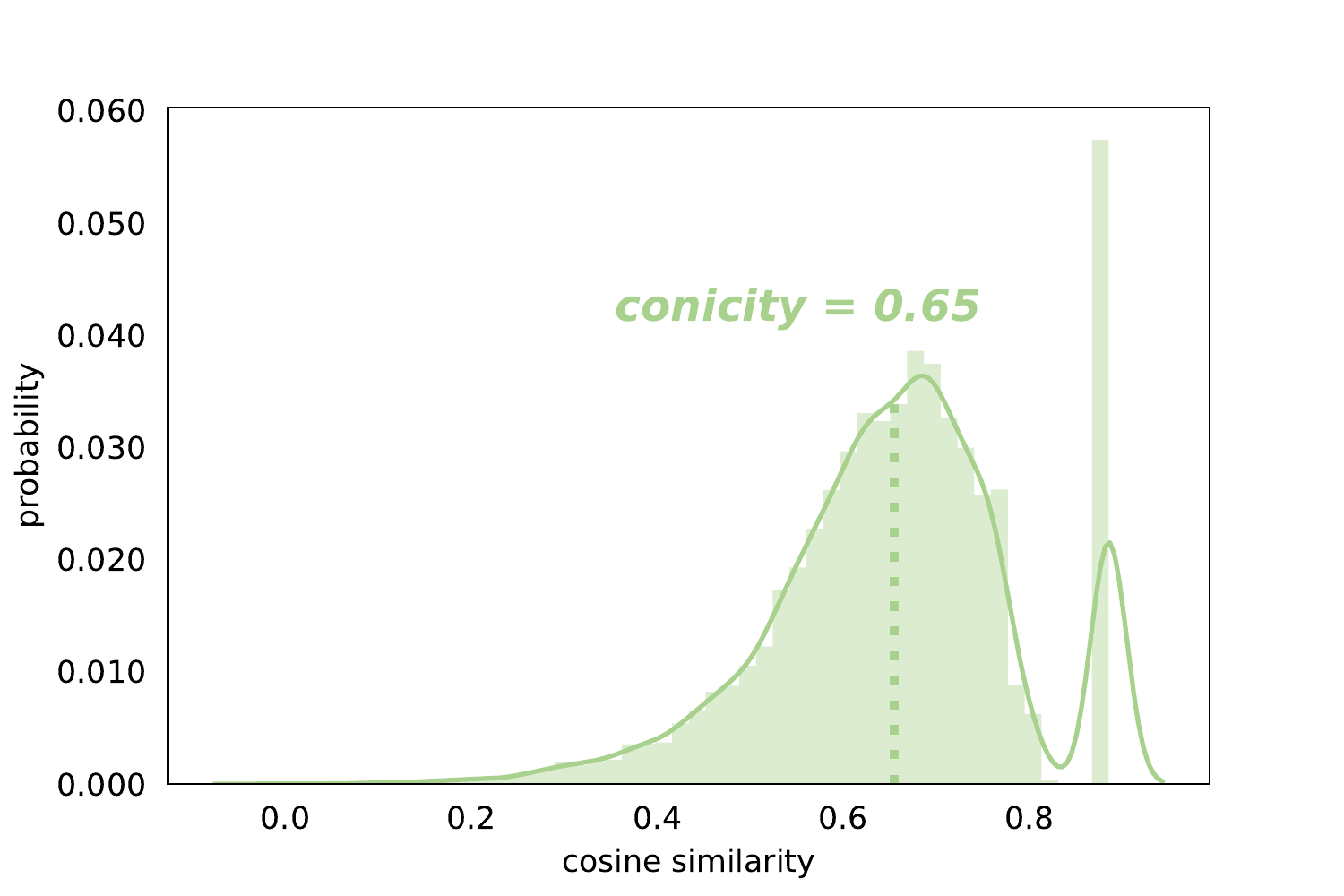}}
		\subfloat[]{\includegraphics[width=0.33\hsize, height = 0.25\hsize]{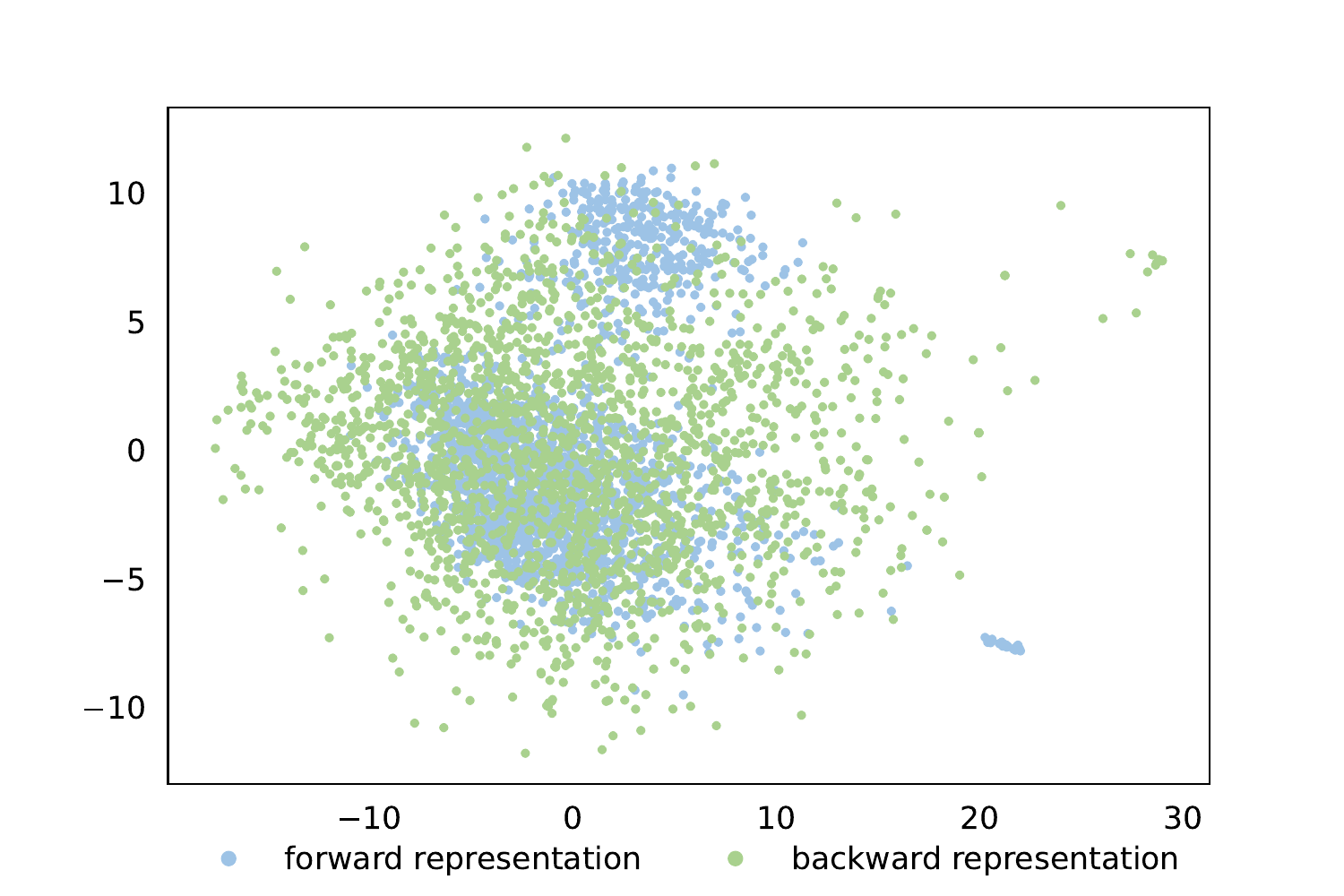}}  
		\quad
		\subfloat[]{\includegraphics[width=0.33\hsize, height = 0.25\hsize]{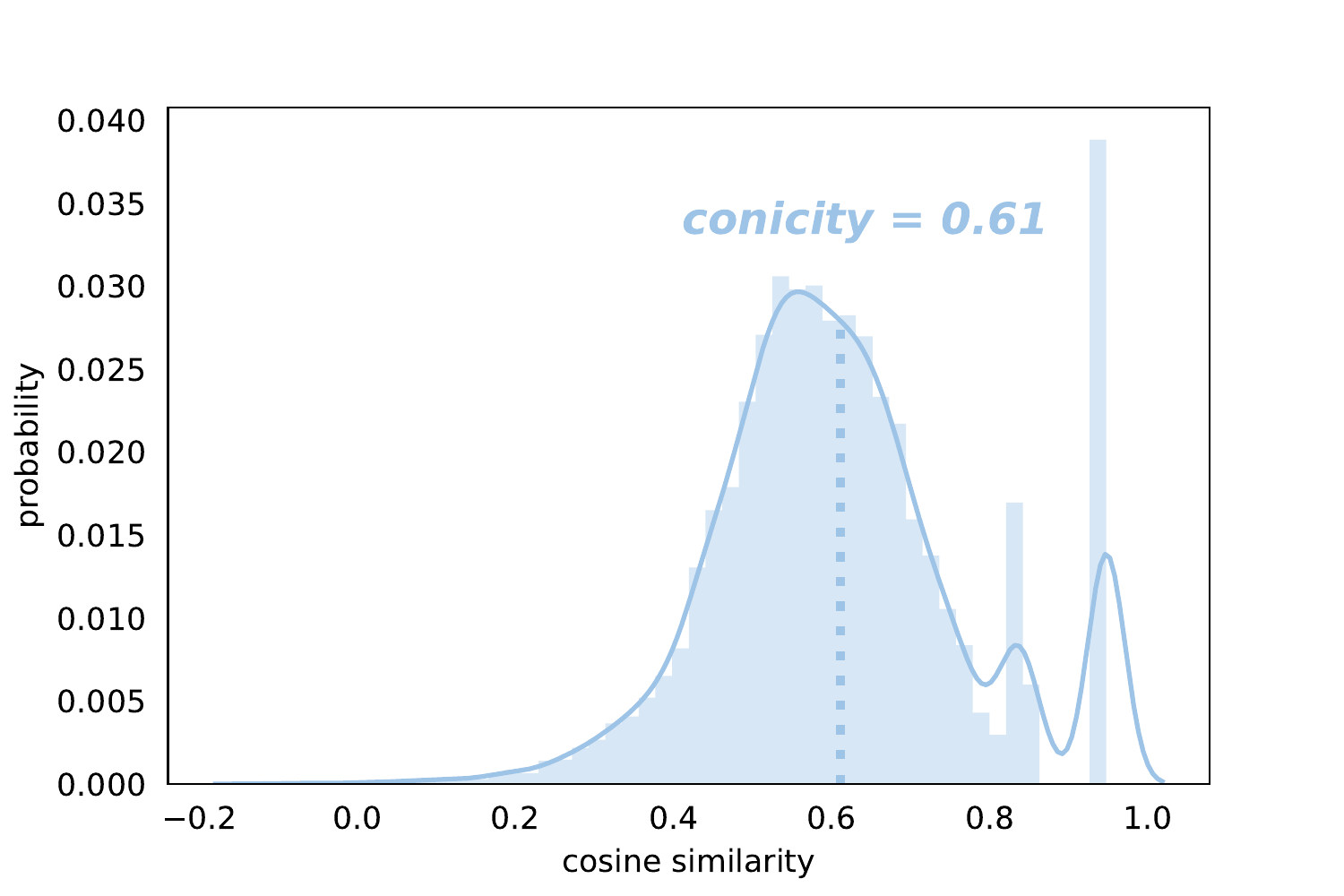}}
		\subfloat[]{\includegraphics[width=0.33\hsize, height = 0.25\hsize]{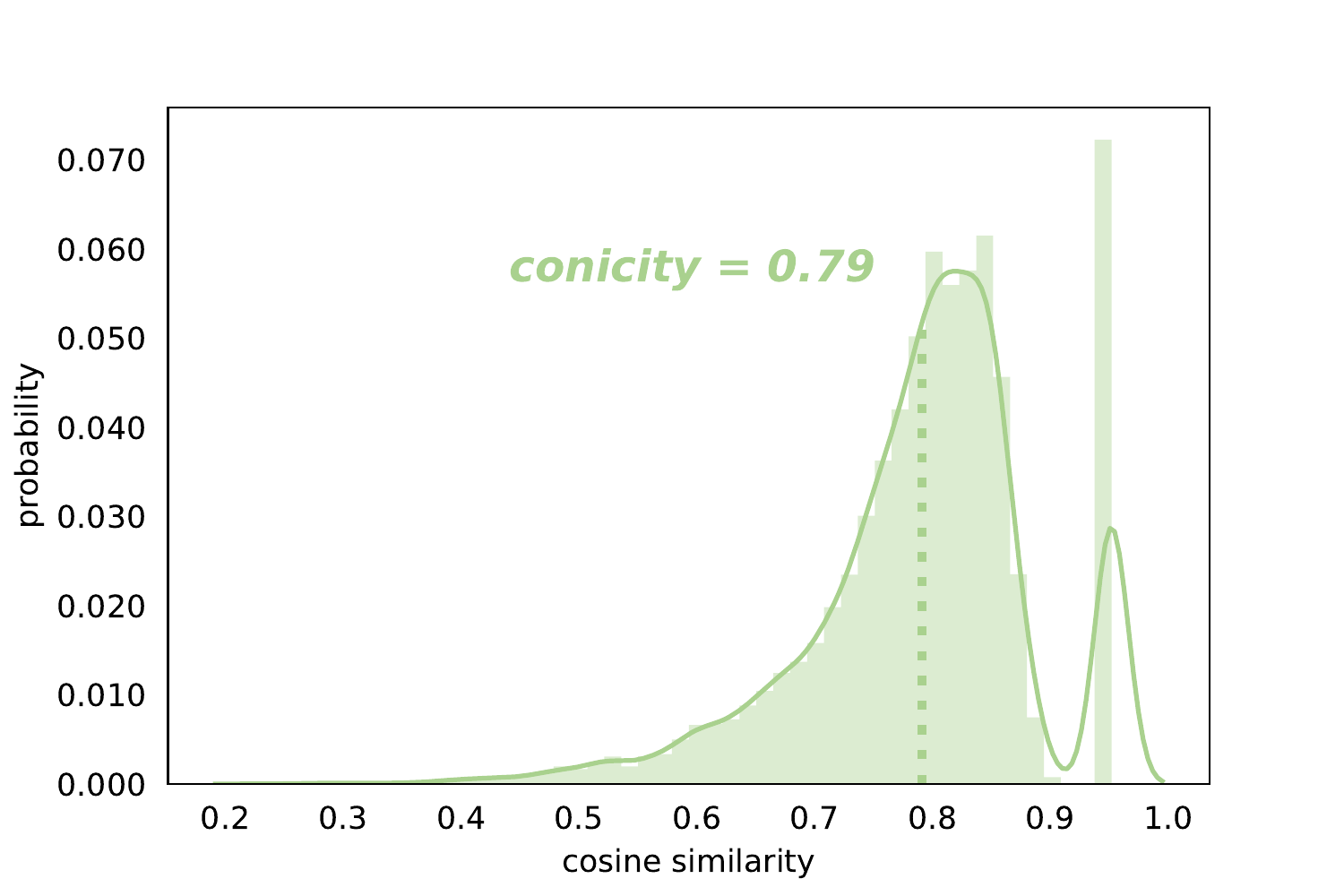}}
		\subfloat[]{\includegraphics[width=0.33\hsize, height = 0.25\hsize]{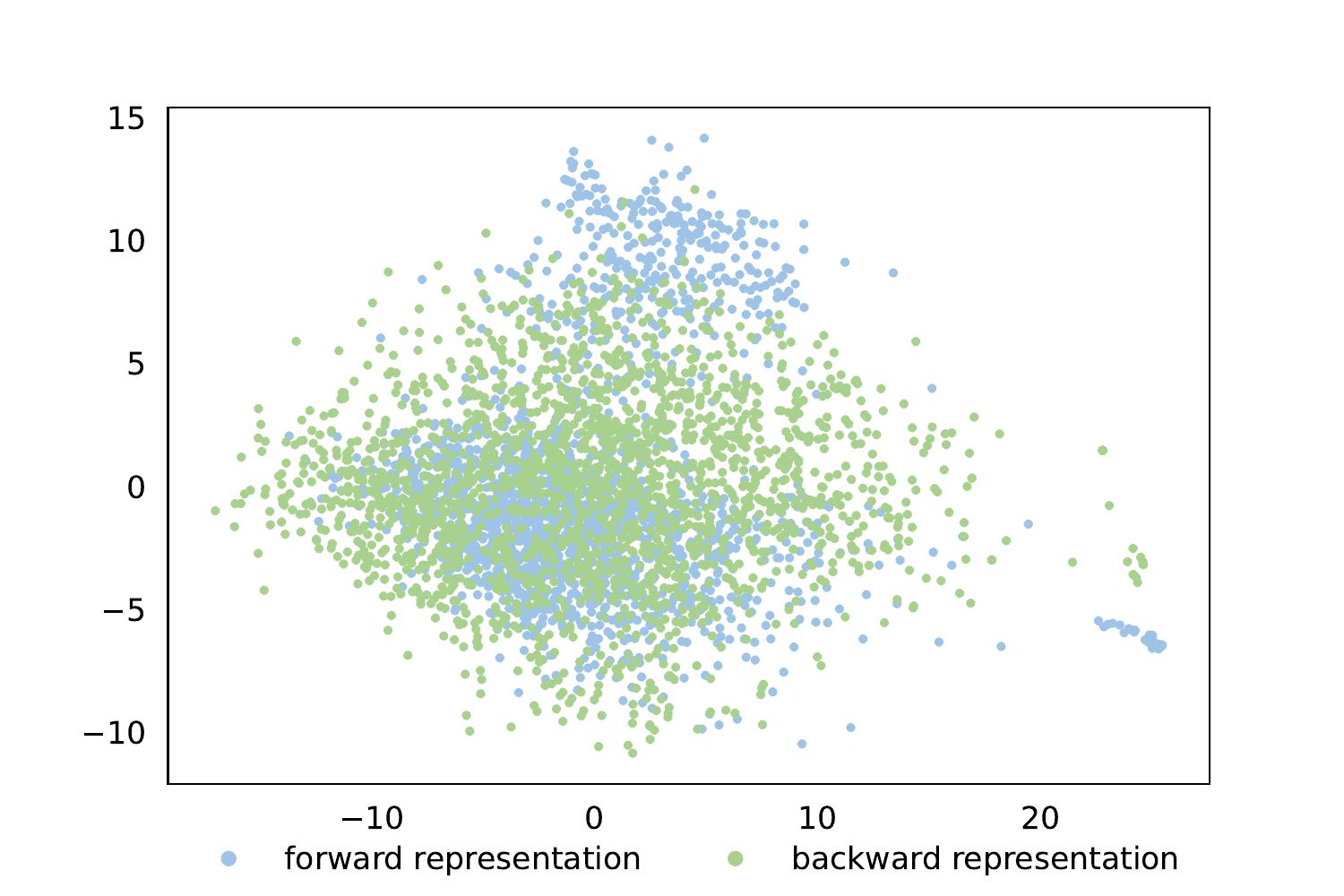}}
		\caption{\emph{The visualization comparison of DsMtGCN and DsMtGCN w/o GC.} (a) The probability density diagram for cosine similarity of $E^f$ from DsMtGCN. (b) The probability density diagram for cosine similarity of $E^b$ from DsMtGCN. (c) The geometric distribution of $E^f$ and $E^b$ from DsMtGCN. (d) The probability density diagram for cosine similarity of $E^f$ from DsMtGCN w/o GC. (e) The probability density diagram for cosine similarity of $E^b$ from DsMtGCN w/o GC. (f) The geometric distribution of $E^f$ and $E^b$ from DsMtGCN w/o GC.}
		\label{gc}
	\end{figure*}
	
	To further investigate the performance of DsMtGCN under different sub-tasks and relation types, we present the results of our model compared with ConvE \cite{conve}, InteractE \cite{interacte}, SACN \cite{sacn} and CompGCN \cite{compgcn} on FB15k-237 in Table \ref{relation_result}. Following previous work, the relations are divided into four categories including one-to-one (1-1), one-to-many (1-N), many-to-one (N-1) and many-to-many (N-N) by counting the average number of tails per head and heads per tail \cite{transh}. It can be observed that both CompGCN and DsMtGCN outperform other models with significant improvements on all four relation types, which can be attributed to the full exploitation of graph structure information. When it comes to DsMtGCN, it gets the highest MRR for all relation types especially complex ones like 1-N, N-1 and N-N, which indicates that aggregating the information from neighbors in different directions specifically can help to handle complex relations under both forward and backward sub-tasks.
	
	\subsection{Ablation Study}
	
	\begin{table}[!t]
		\centering
		\caption{Results of ablation study on FB15k-237.}
		\begin{tabular}{l|cccc}
			\toprule
			\textbf{Model}                     & \textbf{MRR}   & \textbf{H@1} & \textbf{H@3} & \textbf{H@10} \\ \midrule
			DsMtGCN                      & \textbf{0.363} & \textbf{0.273}  & \textbf{0.398}  & \textbf{0.545}   \\ \midrule
			\textit{w/o} GC & 0.360 & 0.268  & 0.397  & 0.542   \\ 
			\textit{w/o} MHSA   & 0.358 & 0.268  & 0.392  & 0.539   \\ 
			\textit{w/o} TU    & 0.361 & 0.271  & 0.396  & 0.542   \\ \bottomrule
		\end{tabular}
		\label{ablation_result}
	\end{table}
	
	Since our model has outperformed other competitors in the aforementioned comparative experiments, to further probe into the contribution of each module to the final performance, we conduct ablation experiments of DsMtGCN on FB15k-237. Considering that the main innovations in this work can be divided into three modules including geometric constraints, multi-head fine-grained self-attention mechanism and label smoothing with triple uncertainty, we abbreviate them as \textbf{GC}, \textbf{MHSA} and \textbf{TU} respectively for the convenience of explanation. In order to analyze the role of each module, we remove them from the original model in turn to get following sub-models: \textbf{DsMtGCN w/o GC}, \textbf{DsMtGCN w/o MHSA} and \textbf{DsMtGCN w/o TU}, where w/o means without specific module or mechanism. Specifically, we set $\lambda_1$ and $\lambda_2$ to 0 to obtain DsMtGCN w/o GC, we replace $A^F$ with the vector filled with 1 to obtain DsMtGCN w/o MHSA, and $k$ is initialized to 0 to get DsMtGCN w/o TU, these sub-models are trained under the same settings, the results are shown in Table~\ref{ablation_result}.
	
	It's obvious that the removal of MHSA brings the greatest decrease in effect, as it enables our model to make full use of structural information from neighbors in different directions at entity level, which is the core of our direction-sensitive multi-task framework. The effect of GC is evident as well, because it can help to improve the geometric distribution of embeddings. Lastly, the introduction of TU proves to be indispensable for further improvements by considering the number of possible tail entities.
	
	In order to show the effect of GC module in a more explicit way, we visualize the cosine similarity and geometric distribution of $E^f$ and $E^b$ in Fig.\ref{gc}, the mean value of cosine similarity (i.e., conicity) is marked by bold dotted line, and the scatter graph shows the distribution of embeddings reduced to two dimensions by principal component analysis (PCA). From Fig.\ref{gc}(a)(b)(d)(e), we can find that curves with GC is more gentle and the center is more left, which means that better diversity is preserved among embeddings. Besides, The points in Fig.\ref{gc}(c) are more evenly distributed than those in Fig.\ref{gc}(f), and the dots with different colors can cover each other more completely in Fig.\ref{gc}(c), indicating the validity of GC module.
	
	\begin{figure*}[!t]
		\centering
		\subfloat[]{\includegraphics[width=0.33\hsize]{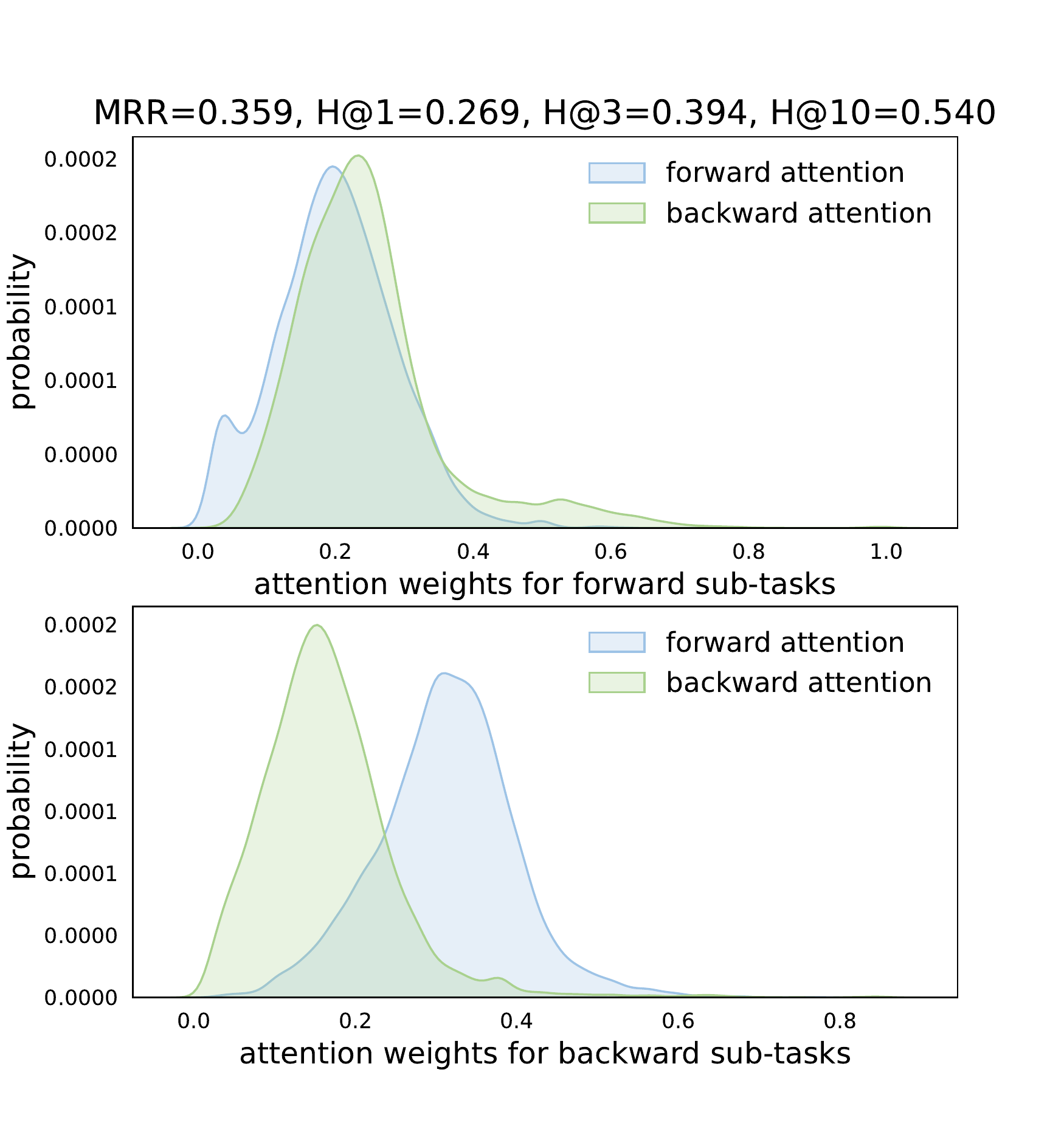}}  
		\subfloat[]{\includegraphics[width=0.33\hsize]{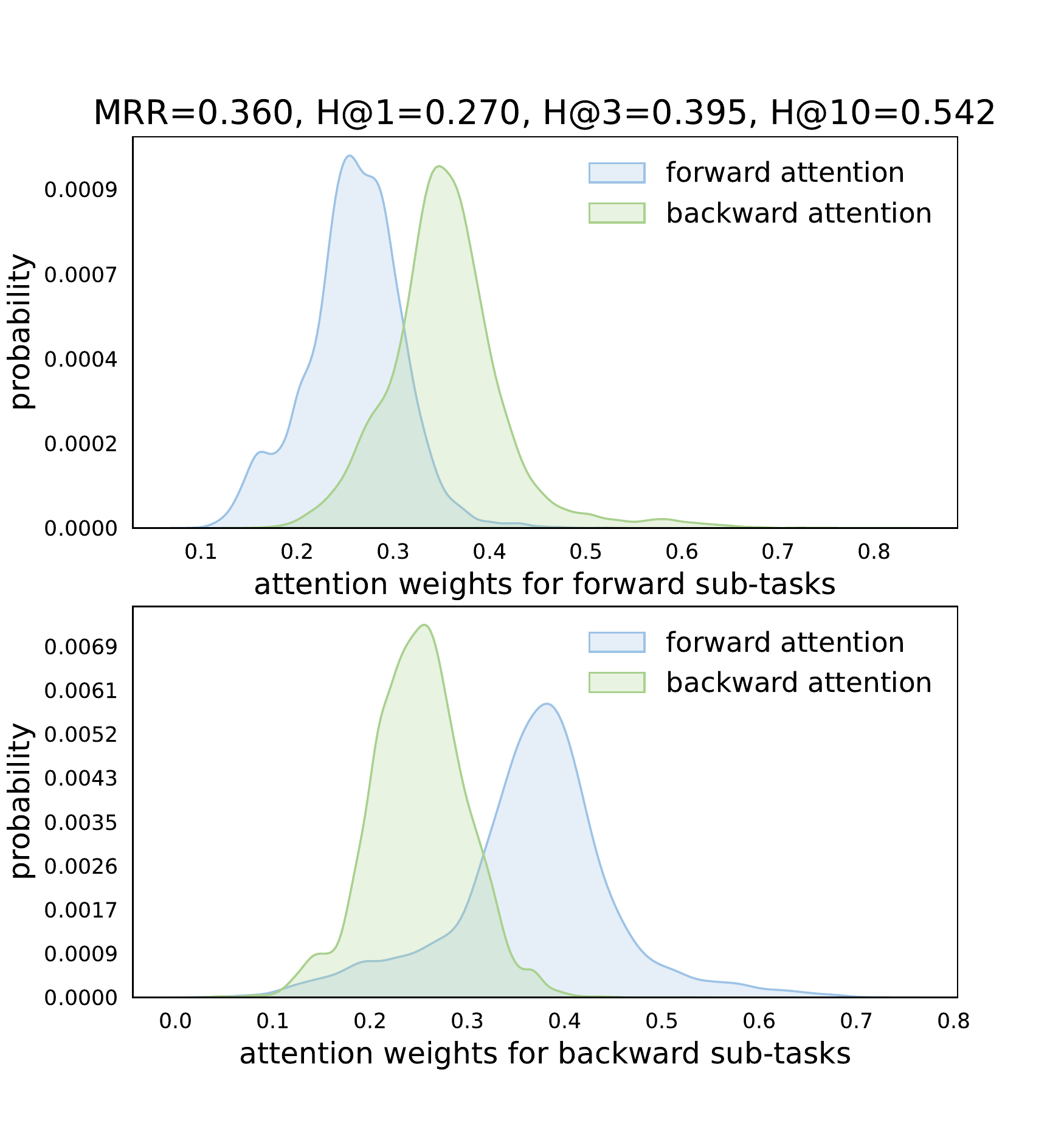}}
		\subfloat[]{\includegraphics[width=0.33\hsize]{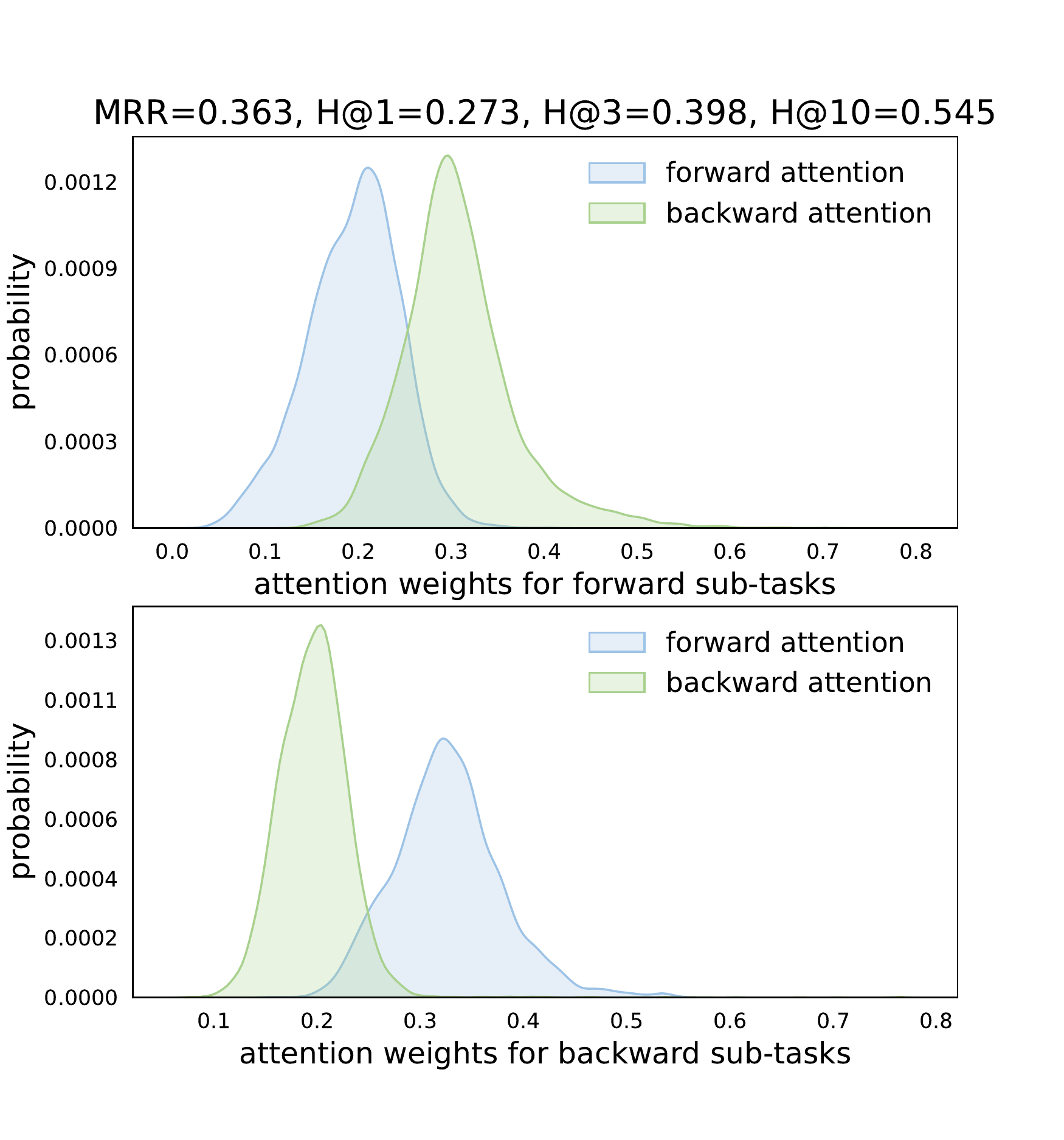}}  
		\caption{\emph{The visualization comparison of DsMtGCN r/w MHAA, DsMtGCN r/w MHPA and DsMtGCN.} (a) The attention distribution and prediction performance for DsMtGCN r/w MHAA. (b) The attention distribution and prediction performance for DsMtGCN r/w MHPA. (c) The attention distribution and prediction performance for DsMtGCN.}
		\label{att}
	\end{figure*}

	\begin{table*}[!t]
		\centering
		\setlength\tabcolsep{4pt}
		\renewcommand\arraystretch{0.1} 
		\caption{Several examples reflecting the preference of forward and backward sub-tasks to neighbors in different directions. The key neighbors playing prompt roles in prediction are bolded, and the relative attention weights of forward and backward neighbors are listed in the rightmost column.}
		\begin{tabular}{l|l|l}
			\toprule
			Task $\Rightarrow$ Target  & Forward $\Vert$ Backward Neighbors  &  Weights \\ \midrule
			\makecell[l]{\textit{(Soul music, artists, ?)} \\ $\Rightarrow$ \textit{Tyrese Gibson}} & \makecell[l]{\textbf{\textit{(artists, Patti Labelle)}}, \textbf{\textit{(artists, Kim Carnes)}} $\Vert$ \\ \textit{(genre$^{-1}$, Funk)}} & (0.66, 0.34) \\ \midrule \midrule
			\makecell[l]{\textit{(John Cale, profession, ?)} \\ $\Rightarrow$ \textit{Composer}} & \makecell[l]{\textit{(profession, Actor)}, \textit{(nationality, Wales)} $\Vert$ \\ \textbf{\textit{(genre$^{-1}$, Art rock)}, \textit{(instrument$^{-1}$, Piano)}}} & (0.29, 0.71) \\ \midrule
			\makecell[l]{\textit{(Who Framed Roger Rabbit, genre, ?)} \\ $\Rightarrow$ \textit{Fantasy}} & \makecell[l]{\textit{(genre, Parody)}, \textit{(genre, Buddy film)}, \\ \textit{(country, USA)}, \textit{(release, DVD)} $\Vert$ \\ \textit{(actor$^{-1}$, Joel Silver)}, \textbf{\textit{(distributor$^{-1}$,Disney)}} \\ \textit{(director$^{-1}$,Robert Zemeckis)} } & (0.32, 0.68) \\ \midrule \midrule
			\makecell[l]{\textit{(Louise Fletcher, nomination$^{-1}$,?)} \\ $\Rightarrow$ \textit{Academy Award for Best Actress}} & \makecell[l]{\textbf{\textit{(profession, Actor)}}, \textbf{\textit{(gender, Femal)}}, \\ \textit{(film, One Flew Over the Cuckoo's Nest)} $\Vert$ \\ \textit{(student$^{-1}$, University of North Carolina)}} & (0.73, 0.27) \\ \midrule
			\makecell[l]{\textit{(Charles S.Dutton, student$^{-1}$, ?)} \\ $\Rightarrow$ \textit{Towson University}} & \makecell[l]{\textbf{\textit{(place\_of\_birth, Baltimore)}}, \\ \textbf{\textit{(place\_live, Baltimore)}}, \\ \textit{(profession, Television director)} $\Vert$ \\ \textit{(student$^{-1}$, Yale University)}, \\ \textit{(ethnicity$^{-1}$, African American)}} & (0.69, 0.31) \\ \midrule \midrule
			\makecell[l]{\textit{(Earth, area$^{-1}$, ?)} \\ $\Rightarrow$ \textit{Egypt}} & \makecell[l]{\textit{(currency, nited States dollar)}, \\ \textit{(taxonomy, Library of Congress Classification)} $\Vert$ \\ \textbf{\textit{(area$^{-1}$, Italy)}}, \textbf{\textit{(area$^{-1}$, Kenya)}}, \\ \textbf{\textit{(area$^{-1}$, Singapore)}}}  & (0.26, 0.74) \\ \bottomrule
		\end{tabular}
		
		\label{case}
	\end{table*}
	
	To further verify the effectiveness of \textbf{MHSA} module, we replace it with multi-head adaptive attention mechanism (\textbf{MHAA}) and multi-head parameter-generated attention mechanism (\textbf{MH-PA}). In particular, \textbf{MHAA} implements attention weights with randomly initialized learnable vectors, \textbf{MHPA} applies linear transformation from the embedding matrix to attention weights, the attention distribution generated by above models and corresponding prediction effects are shown in Fig.~\ref{att}, \textit{r/w} means to replace \textbf{MHSA} with above substitutes. It's clear that while dealing with different sub-tasks, the preference for neighbor information in different directions changes evidently, the backward neighbors are paid more attention under the forward sub-tasks, while the opposite happens for backward sub-tasks. Besides, there exists intersections between curves, which means that even for the same sub-tasks, different entities have different preferences, demonstrating the importance of adopting attention at entity level. Ultimately, the prediction effect in the graph increases gradually from left to right, and the overlapping parts between different curves become smaller, indicating that the \textbf{MHSA} module can better allocate attention weights according to different sub-tasks and entities.

	\subsection{Case Study}	
	We employ case study to further analyze why DsMtGCN works, several examples are shown in Table \ref{case}, where tasks and corresponding targets come from $\mathcal{T}^{test}$, forward and backward neighbors are taken from $\mathcal{T}^{train}$. Generally speaking, backward neighbors can provide more information to forward sub-tasks, and forward neighbors are more beneficial to backward sub-tasks, because they can provide more complementary information in most cases. For example, while dealing with \textit{(Who Framed Roger Rabbit, genre, ?)}, the neighbors in same direction like \textit{(genre, Parody)}, \textit{(genre, Buddy film)} are unable to provide additional information about \textit{Fantasy} as they are at the same level, while \textit{(distributor$^{-1}$,Disney)} gives a strong signal; when the query is \textit{(Charles S.Dutton, student$^{-1}$, ?)}, knowing \textit{Charles S.Dutton} is a student of \textit{Yale University} is helpless, but his deep connection with \textit{Baltimore} can remind us of the \textit{Towson University} in \textit{Baltimore}. On the other hand, this rule is not true for queries about abstract concepts like \textit{(Soul music, artists, ?)} or \textit{(Earth, area$^{-1}$, ?)}, as the opposite direction usually means abstract hierarchy, while the answer is concrete.
	
	\section{Conclusion}
	\label{cc}
	
	In this work, we notice that the neighbor information aggregated in forward and backward directions has specific significance for different entities and sub-tasks. For taking advantage of such sensitivity to direction, we propose a novel multi-task GCN for knowledge graph, DsMtGCN, which decomposes the single link prediction tasks into forward sub-tasks as well as backward sub-tasks, the multi-head self-attention is utilized to combine neighbor information from different directions, the geometric constraints and label smoothing with triple uncertainty are applied to obtain further performance gains. Through extensive comparison and ablation experiments, we demonstrate the effectiveness and rationality of our model. For future work, we will explore richer interactions between entities and relations, the idea of triple uncertainty is also expected to be applied in more multi-label classification tasks.
	
	\bibliography{mybibfile}
	
\end{document}